\documentclass[cameraready]{Interspeech}

\title{Vimarsha: Faithful ASR Evaluation for Indian Languages with Demographic Diversity, In-the-Wild Audio and Spelling Variations}

\author[affiliation={1}, orcid=0009-0004-7409-0034]{Kaushal}{Bhogale}
\author[affiliation={1}]{Srija}{Anand}
\author[affiliation={2}]{Sadakopa Ramakrishnan}{Thothathiri}
\author[affiliation={1}]{Tahir}{Javed}
\author[affiliation={2}]{Sshubam}{Verma}
\author[affiliation={1}]{Mitesh M.}{Khapra}

\address{
    $^1$ Indian Institute of Technology, Madras, India \quad
    $^2$ Sarvam AI, India
}
\email{cs22d006@cse.iitm.ac.in, miteshk@dsai.iitm.ac.in}


\usepackage{comment}
\usepackage{multirow}
\usepackage[normalem]{ulem}

\begin{document}

\maketitle

\begin{abstract}
Evaluation benchmarks for Indian language automatic speech recognition (ASR) suffer from two systematic biases: optimistic scores from clean, controlled audio conditions, and pessimistic scores from overly rigid transcription standards that penalize valid linguistic variations. We introduce Vimarsha, a 100-hour benchmark spanning all 22 scheduled Indian languages, designed to address both distortions. Vimarsha combines demographically diverse on-field recordings with carefully mined in-the-wild audio selected for acoustic difficulty, alongside a lattice of variations framework that encodes multiple valid transcriptions per utterance. Evaluations of 10 state-of-the-art ASR models reveal substantial shifts in model rankings under realistic conditions, geographic and demographic performance disparities, and systematic failure modes across speaking rates and acoustic environments.
\end{abstract}

\section{Introduction}

\begin{figure}[t]
    \centering
    \includegraphics[width=\linewidth]{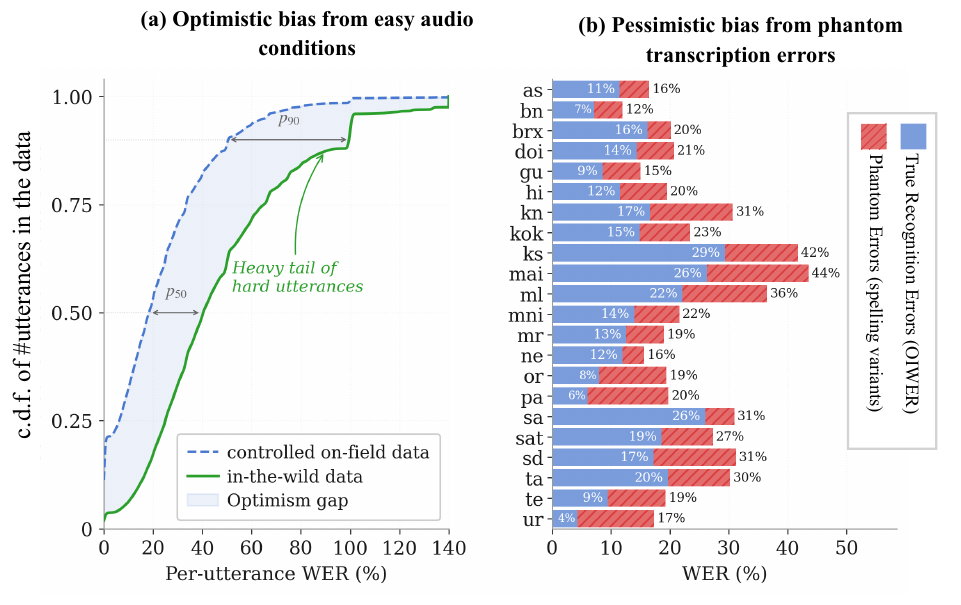}
    \caption{Two evaluation biases in Indian language ASR benchmarks. \textbf{(a)} Easy audio conditions yield optimistic WER by sampling from the low-error tail; shaded region shows the gap relative to Vimarsha. \textbf{(b)} Rigid transcription inflates WER with \textit{phantom errors} (penalizing for spelling variants).}
    \label{fig:evaluation-biases}
\end{figure}

India presents one of the most linguistically diverse settings for speech recognition, with 22 constitutionally recognized languages, and speakers operating in multilingual environments. Recent years have witnessed rapid progress in ASR for Indian languages, driven by the availability of large multilingual datasets \cite{indicvoices, springinx,10096933,DBLP:conf/aaai/JavedDRBRKKK22,indicsuperb} and increasingly capable models \cite{mahadhwani,vistaar,DBLP:conf/aaai/JavedDRBRKKK22,indicsuperb}. However, as model development accelerates, reliable evaluation has not kept pace. We argue that current evaluation benchmarks for Indian language ASR can be systematically misleading in two orthogonal ways, as shown in Figure \ref{fig:evaluation-biases} and discussed below.

The first issue relates to the lack of acoustic realism in existing benchmarks. Real-world deployment environments involve background noise, channel distortions, large demographic diversity, and spontaneous speech. These conditions are largely absent from existing evaluation corpora, which are predominantly curated under clean, controlled or simulated settings \cite{indicvoices, lahaja, svarah}. This creates an \textit{optimistic} bias: benchmark scores that do not generalize to the messy real world acoustic conditions.

The second limitation is the mirror image of the first. While benchmark audio is often too easy, transcription standards are overly rigid. Indian languages exhibit fluid spelling conventions and non-standard spellings for code mixed English words. Enforcing a single canonical reference penalizes valid recognitions and introduces a \textit{pessimistic} bias. Although prior work has examined this issue for English and Indian languages \cite{ali2015,ali2019variability,karita2023,manohar2023morphological,srivastava2018,snwer}, a benchmark that simultaneously captures real world acoustic difficulty and supports multiple valid transcripts remains missing. 

This paper introduces \textit{Vimarsha}, a benchmark addressing both distortions. First, we construct a 100-hour evaluation set spanning 22 Indian languages from two complementary sources: an on-field collected dataset for demographic and speaker diversity, and a carefully mined in-the-wild corpus with real-world acoustic conditions. Rather than sampling in-the-wild audio uniformly, we select hard segments, ensuring the benchmark challenges model capability. Second, we replace single reference transcription with a \textit{lattice of variations} framework: a principled and practical ground truth representation that captures the acceptable space of correct transcriptions for each utterance, including spelling variants, code mixed forms, and phonetically equivalent forms. This lattice is built using a human annotation pipeline and enables evaluation against functional correctness rather than orthographic conformity.

We evaluate 10 state-of-the-art ASR models against Vimarsha and demonstrate that model rankings shift substantially depending on acoustic condition and reference flexibility. These findings that have direct implications for how model comparisons are reported and interpreted. The benchmark, evaluation protocols, and model results are released openly\footnote{\url{https://github.com/AI4Bharat/Vimarsha}}.

\begin{table*}[h!]
\centering
\small
\setlength{\tabcolsep}{3pt}
\renewcommand{\arraystretch}{1.05}
\caption{Per-language statistics of the Vimarsha benchmark}
\label{tab:benchmark_stats_transposed}

\resizebox{\textwidth}{!}{
\begin{tabular}{llcccccccccccccccccccccc|c}
\toprule
\textbf{Metric} & \textbf{Split}
& \textbf{as} & \textbf{bn} & \textbf{brx} & \textbf{doi} & \textbf{gu} & \textbf{hi} & \textbf{kn} & \textbf{ks} & \textbf{kok} & \textbf{mai} & \textbf{ml} & \textbf{mni} & \textbf{mr} & \textbf{ne} & \textbf{or} & \textbf{pa} & \textbf{sa} & \textbf{sat} & \textbf{sd} & \textbf{ta} & \textbf{te} & \textbf{ur} & \textbf{Total} \\
\midrule

\multirow{3}{*}{\textbf{Hours}}
& \textbf{COF}
& 3.2 & 2.0 & 2.0 & 0.8 & 2.7 & 1.9 & 2.1 & 1.1 & 0.7 & 1.8 & 1.6 & 2.7 & 2.4 & 3.8 & 2.0 & 1.8 & 0.9 & 1.7 & 0.7 & 2.8 & 2.8 & 1.8 & \textbf{43.3} \\

& \textbf{IW}
& 3.6 & 2.7 & 2.4 & 0.0 & 5.4 & 3.0 & 3.5 & 1.2 & 1.2 & 2.9 & 2.6 & 3.1 & 3.2 & 3.0 & 2.5 & 4.6 & 2.0 & 1.0 & 0.4 & 2.6 & 2.9 & 2.5 & \textbf{56.1} \\

& \textbf{Total}
& 6.7 & 4.7 & 4.3 & 0.8 & 8.1 & 4.9 & 5.6 & 2.3 & 1.9 & 4.7 & 4.3 & 5.8 & 5.6 & 6.8 & 4.5 & 6.5 & 3.0 & 2.7 & 1.1 & 5.4 & 5.6 & 4.3 & \textbf{99.4} \\

\midrule

\multirow{2}{*}{\textbf{Vocab (k)}}
& \textbf{COF}
& 9.9 & 6.6 & 6.4 & 3.1 & 9.8 & 6.4 & 10.5 & 5.5 & 3.4 & 8.5 & 9.2 & 9.2 & 8.6 & 10.8 & 7.8 & 6.3 & 3.9 & 6.1 & 3.6 & 12.8 & 11.2 & 6.4 & \textbf{166} \\

& \textbf{IW}
& 11.4 & 9.0 & 7.4 & --- & 17.4 & 9.8 & 19.1 & 7.3 & 7.1 & 12.5 & 15.0 & 14.4 & 12.4 & 10.4 & 11.5 & 11.2 & 8.0 & 4.6 & 2.3 & 13.9 & 13.4 & 7.2 & \textbf{225} \\

\midrule

\multirow{2}{*}{\textbf{SNR (dB)}}
& \textbf{COF}
& 38.2 & 44.4 & 28.4 & 30.9 & 34.2 & 47.0 & 37.6 & 42.2 & 40.4 & 39.2 & 44.7 & 40.0 & 41.1 & 38.7 & 40.3 & 43.4 & 34.6 & 35.1 & 39.1 & 39.6 & 37.3 & 34.5 & \textbf{38.7} \\

& \textbf{IW}
& 9.8 & 8.7 & 6.8 & --- & 10.5 & 7.1 & 8.0 & 12.4 & 12.1 & 8.6 & 11.3 & 13.0 & 7.5 & 9.3 & 9.4 & 8.9 & 20.5 & 13.2 & 14.5 & 11.1 & 8.7 & 15.8 & \textbf{10.8} \\

\midrule

\textbf{avg\_var/word}
&
& 1.4 & 1.4 & 1.3 & 1.6 & 1.5 & 1.5 & 1.9 & 1.9 & 1.7 & 1.9 & 1.8 & 1.5 & 1.4 & 1.3 & 1.7 & 1.4 & 1.6 & 1.5 & 1.6 & 1.7 & 1.7 & 1.4 & \textbf{1.5} \\

\midrule

\textbf{\#districts}
& \textbf{COF}
& 20 & 21 & 4 & 5 & 13 & 36 & 22 & 10 & 7 & 11 & 14 & 11 & 29 & 4 & 24 & 20 & 65 & 12 & 8 & 37 & 40 & 29 & \textbf{356} \\

\textbf{\#speakers}
& \textbf{COF}
& 472 & 274 & 131 & 65 & 150 & 427 & 209 & 148 & 83 & 242 & 311 & 176 & 317 & 200 & 230 & 245 & 135 & 143 & 140 & 476 & 257 & 268 & \textbf{5099} \\

\bottomrule
\end{tabular}}

\end{table*}

\section{Related Work}

\textbf{Benchmarks for Indian Languages} Early resources such as MUCS 2021 \cite{mucs} and OpenSLR corpora \cite{openslr1, openslr2, googletts} covered few languages under read speech. IndicSUPERB \cite{indicsuperb} introduced Kathbath across 12 languages and tasks, while Vistaar \cite{vistaar} aggregated 59 benchmarks and showed strong variation in model rankings across evaluation sets. IndicVoices \cite{indicvoices} scaled coverage to all 22 scheduled languages, and SPRING-INX \cite{springinx} and Svarah \cite{svarah} addressed conversational and accented English speech. However, most of these benchmarks rely on controlled recordings and standardized transcripts. 

\noindent\textbf{Challenging and Noise-Robust ASR Evaluation} The CHiME series \cite{barker2015chime3} highlighted performance degradation under real world noise, while GigaSpeech 2 \cite{yang2024gigaspeech2} extended such evaluation using naturally noisy web audio. In Indian languages, Kathbath-Hard \cite{indicsuperb} simulates robustness via noise augmentation rather than natural challenging recordings.

\noindent\textbf{Variation-Aware Evaluation} Standard WER penalizes acceptable spelling variation. Multi reference WER \cite{ali2015,ali2019variability} and lattice based CER \cite{karita2023} address this for Arabic and Japanese. For Indian languages, WER over penalizes valid morphological variations \cite{manohar2023morphological}, while phoneme based metrics \cite{srivastava2018,snwer} rely on incomplete G2P resources. OIWER \cite{bhogale2025oiwer} generates orthographic variation sets using LLMs, reducing pessimistic WER inflation by 6.3 points across 22 languages. Vimarsha extends this framework while additionally addressing acoustic realism.

\section{The Vimarsha Benchmark}

\subsection{Data Collection Procedure}

\textbf{Controlled On-Field (COF) Data} Our on-field data collection followed the protocol established by IndicVoices \cite{indicvoices}. Specifically, we recruited native speakers from diverse districts, covering all 22 languages. We ensured that participants were stratified across key demographic attributes, including age, gender, occupation, education, and region. We elicited three types of speech: (i) read speech, in which speakers read predefined sentences; (ii) extempore speech, in which speakers recorded spontaneous responses to prompts; and (iii) conversational speech, involving dialogue between two speakers on a given topic (collected over a telephony channel at 8 kHz). 
To ensure content diversity, we followed the same prompt design criteria for read, extempore, and conversational speech as in \cite{indicvoices}. For quality control, we annotated each audio sample across 7 categories following \cite{indicvoices}. The collected dataset contains 43.3 hours from 5100 speakers across 356 districts covering all 22 languages.

\noindent\textbf{In-the-Wild (IW) Audio Data} To complement controlled recordings, we curate naturally occurring speech from public internet videos, capturing realistic conditions such as background noise, channel distortions, and spontaneous speech. Collection proceeds in two stages: broad sourcing followed by targeted selection of acoustically challenging samples.

\noindent\textbf{Audio Sourcing} We begin with a manually curated seed list of domains and sub domains such as news, education, entertainment, and politics. Using the Serper API, we retrieve relevant websites and extract domain entities with an LLM, which are then used to generate YouTube search queries. Using these queries, we identify channels through the YouTube API, collect their videos, and segment the audio into evaluation ready clips using VAD based chunking.

\noindent\textbf{Selecting Hard Audios} From the collected chunks, acoustically challenging samples are identified using two signals. Each chunk is transcribed by three independently trained ASR models, and samples with high inter-model disagreement (measured via CER) are flagged as difficult. We also run BEATs~\cite{Chen2022beats} as a paralinguistic classifier to detect adverse acoustic conditions such as noise, laughter, or other non-speech events. Chunks satisfying either criterion are retained as hard samples, then uniformly sampled across 86 acoustic tags to maintain diversity.

\subsection{Lattice of Variations}
\label{sec:lattice}

\begin{figure*}[h!]
\centering
\begin{minipage}{0.65\textwidth}
    \centering
    \includegraphics[width=\linewidth]{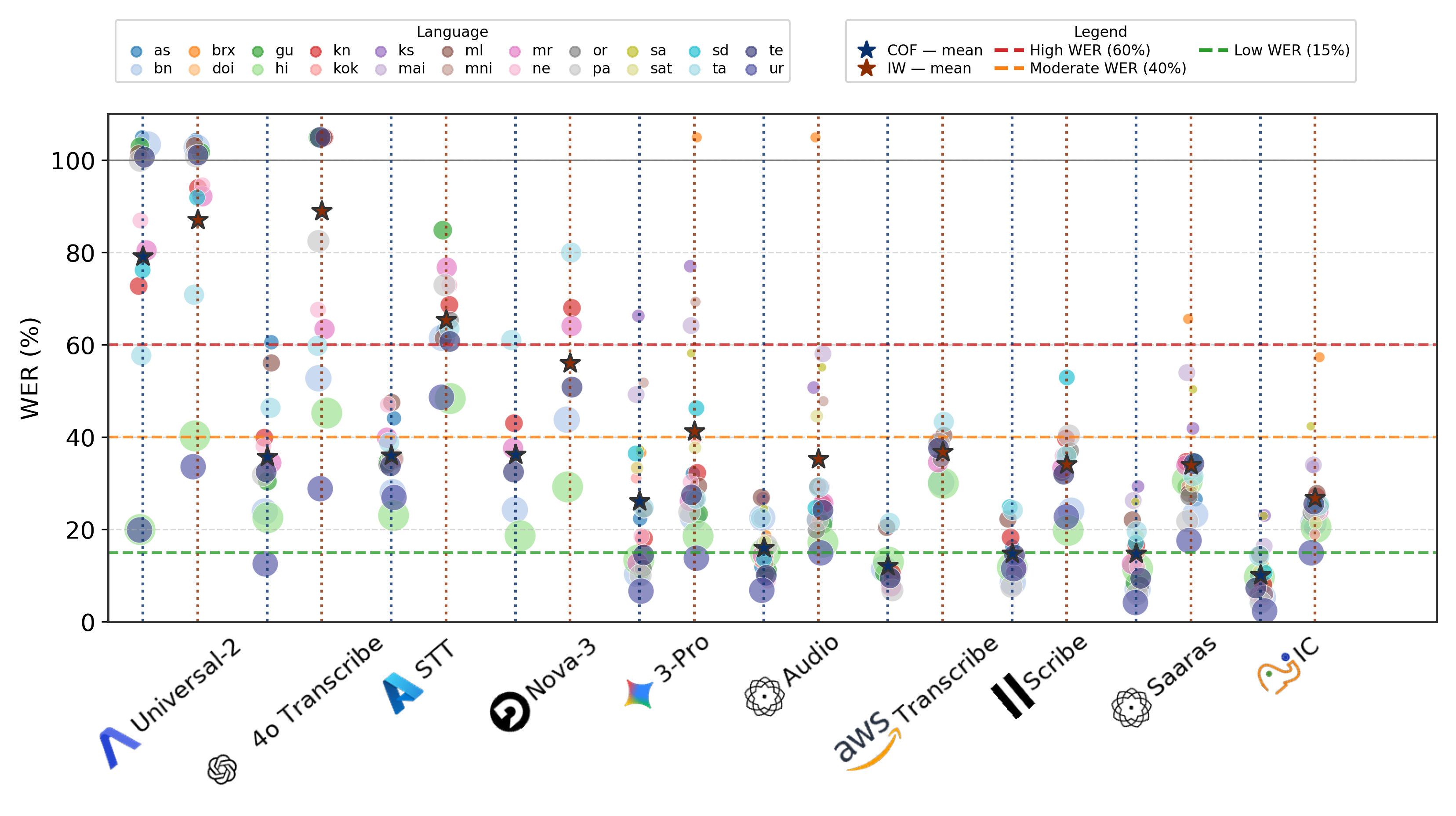}
\end{minipage}
\hfill
\begin{minipage}{0.3\textwidth}
    \centering
    \includegraphics[width=\linewidth]{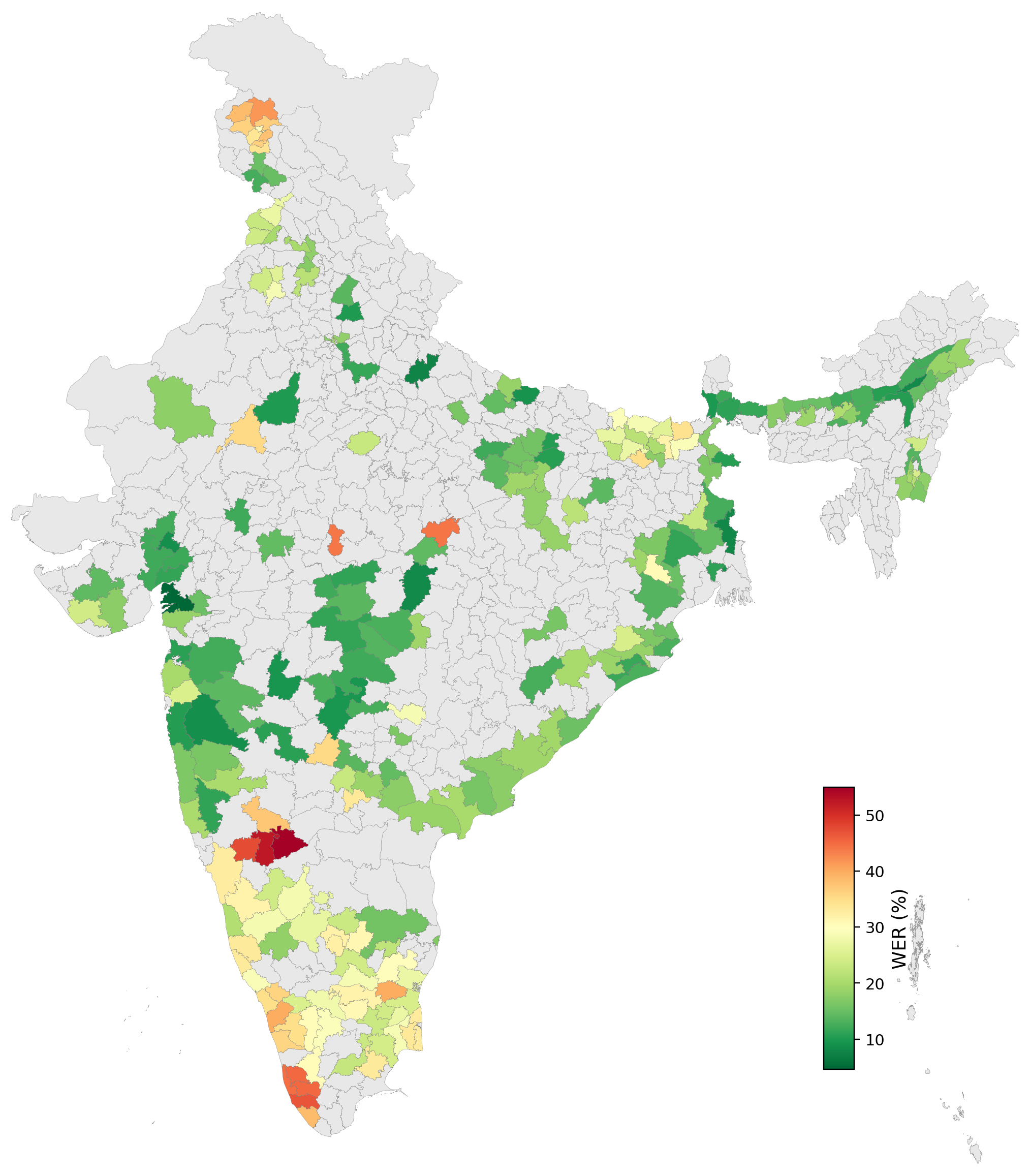}
\end{minipage}

\caption{
\textbf{Left:} WER by model and split (COF and IW). Point size reflects language speaking population; stars indicate mean WER. Values $>$ 100\% are plotted above the 100\% line.
\textbf{Right:} Geographic distribution of ASR performance across Indian districts (WER).
}
\label{fig:main_and_map}
\end{figure*}

\textbf{Motivation.} Indian languages exhibit substantial orthographic variability. Loanwords rarely have standardized spellings, diacritics and compound boundaries are written inconsistently, and code-mixed expressions are rendered phonetically in native scripts, yielding multiple valid spellings of the same term. Standard WER, which evaluates against a single fixed reference, penalizes all such variants equally, producing systematically pessimistic estimates of ASR performance.

Capturing permissible variants is itself challenging. OIWER~\cite{bhogale2025oiwer} proposes generating alternatives via LLMs, but this risks hallucinated forms. Instead, we derive candidates from the outputs of multiple strong ASR models on the same utterance, naturally surfacing confusable spellings that occur in practice. These candidates are then reviewed by human annotators, who validate, add, or remove alternatives based on linguistic knowledge. The result is a structured \textit{lattice of variations} enabling evaluation that better reflects real transcription variability.

\noindent\textbf{Lattice construction.}
Let $\mathcal{M} = \{m_1, \ldots, m_K\}$ be a set of $K$ ASR models. Given a reference utterance, we obtain $K$ hypotheses and align them incrementally: beginning with a pairwise alignment, we merge remaining hypotheses one by one, anchoring on common tokens and grouping divergent spans into variant sets. The resulting lattice $\mathcal{L} = \langle G_1, G_2, \ldots, G_N \rangle$ is a sequence of \textit{variant groups}, where each $G_i \subseteq \Sigma^+$ contains acceptable surface forms at position $i$. WER is computed against the best-matching path through the lattice rather than a single fixed reference, avoiding both the under-permissiveness of single-reference evaluation and the over-permissiveness of exhaustive rule-based expansion.

\noindent\textbf{Human Verification and Augmentation.}
The model-derived lattice serves as an initial scaffold for annotation. Annotators followed per-language guidelines based on the OIWER taxonomy~\cite{bhogale2025oiwer}, covering variation types including matra, diacritic changes, loanword spellings, compound splitting or merging, phonetic spellings, ligature variants, sandhi alternations, and inverse text normalization forms. We recruited 133 annotators across 22 languages, organized as makers and supercheckers. Makers reviewed each lattice to validate, remove, or add variants; supercheckers verified the final set. Annotators could supplement variants using web search as usage evidence, ensuring lattices reflect realistic and lingustically valid ASR outputs.

\subsection{Vimarsha Benchmark Statistics}

Table \ref{tab:benchmark_stats_transposed} covers approximately 100 hours of transcribed speech across all 22 scheduled Indian languages, comprising two subsets: Controlled On-Field \textit{COF} (43.3 hours) from on-field recordings spanning 356 districts and 5{,}099 speakers, and In-the-Wild \textit{IW} (56.1 hours) sourced across 86 diverse acoustic tags, with per-language durations ranging from 0.8 hours (Dogri) to 8.1 hours (Gujarati). The subsets differ sharply in acoustic conditions, with COF averaging 38.7\,dB SNR compared to 10.8\,dB for IW, while vocabulary ranges from 3.1K to 19.1K unique word types across languages. Lattice-based transcription encodes orthographic variation, yielding 1.5 acceptable variants per word on average. 

\section{Evaluation Setup}

\subsection{Models Evaluated}
We evaluate 10 ASR systems supporting 7 - 22 scheduled Indian languages, grouped into India-centric and global models. \textbf{India-centric models} include IndicConformer (I-Conf) \cite{indicvoices}, AI4Bharat's open-source multilingual model supporting all 22 languages; Saaras \cite{saaras}, Sarvam AI's commercial API with strength in noisy and mixed-language speech; and Sarvam Audio \cite{sarvamaudio}, an audio language model built on Sarvam-3B, both supporting all 22 languages. \textbf{Global commercial APIs} include AWS Transcribe (10 languages), Azure STT (13 languages), AssemblyAI Universal-2 (13 languages), ElevenLabs Scribe v2 (14 languages), GPT-4o Transcribe (15 languages), Deepgram Nova-3 (6 languages), and Gemini 3 Pro \cite{gemini} (22 languages). All models are evaluated via their respective batch or streaming APIs under default configurations, with language and script-specific prompting applied where supported.

\subsection{Evaluation Metric}

Following \cite{bhogale2025oiwer}, we evaluate all systems using Orthographically-Informed Word Error Rate (OIWER), which computes the minimum edit distance between a model prediction and a structured reference that encodes permissible orthographic alternatives. In our setting, this reference corresponds directly to the lattice of variations $\mathcal{L} = \langle G_1, G_2, \ldots, G_N \rangle$ constructed in Section~\ref{sec:lattice}, where each variant group $G_i$ defines the set of acceptable surface forms at position $i$. OIWER thus reduces to WER computed against the best-matching path through the lattice rather than a single fixed reference, correcting for the systematic pessimism introduced by fluid spelling conventions and multiple orthographic realizations of loanwords in Indian languages.

\section{Results and Discussion}

\subsection{Evaluating models on the Vimarsha Benchmark}


As shown in Table~\ref{tab:benchmark_stats_transposed}, the COF and IW splits reveal a clear domain gap with consistent re-rankings across models. Every model degrades on the IW split, with the best-performing systems (I-Conf, Saaras) rising from approximately 10--15\% to 27--34\% average WER, confirming that the IW split is a substantially harder evaluation condition. I-Conf leads on both splits, but the relative ordering below it shifts: AWS Transcribe, competitive on COF (12.2), degrades most sharply on IW (36.8, a 202\% relative increase), while Sarvam Audio and Saaras hold their positions more stably, suggesting these systems are more robust to acoustic variation. The most striking finding is \texttt{GPT-4o Transcribe}: competitive on COF (35.8) but near-failure on IW (89.0), with individual language WERs reaching 167.2 on Assamese and 154.4 on Kannada. 

At the language level, Bodo and Kashmiri are universally difficult across both splits and all models, while Manipuri exhibits a large inter-model gap on COF itself (I-Conf: 10.6 vs.\ Gemini: 51.8), decoupling script and morphological difficulty from acoustic conditions. Together, these results demonstrate that COF performance alone is insufficient for judging deployment readiness, and that the IW split is a critical stress test for robustness of multilingual ASR to real-world speech.

\subsection{Does speaker demographics affect ASR performance?}

Across all demographic dimensions, I-Conf consistently achieves the lowest WER, with Saaras and Sarvam Audio close behind, and Gemini Pro trailing by a large margin. Scenario type has the strongest effect overall. All models degrade substantially on conversational speech, with Gemini Pro reaching nearly 39\% WER compared to 22\% on read speech. Notably, I-Conf shows very little sensitivity to any demographic factor, varying by less than 2 WER points across age, area, and gender. In contrast, Gemini Pro is disproportionately affected by speaker qualification level, with a 5+ point gap between speakers with no schooling and graduates, a disparity largely absent in the other models.

\begin{figure}[t]
    \centering
    \includegraphics[width=\linewidth]{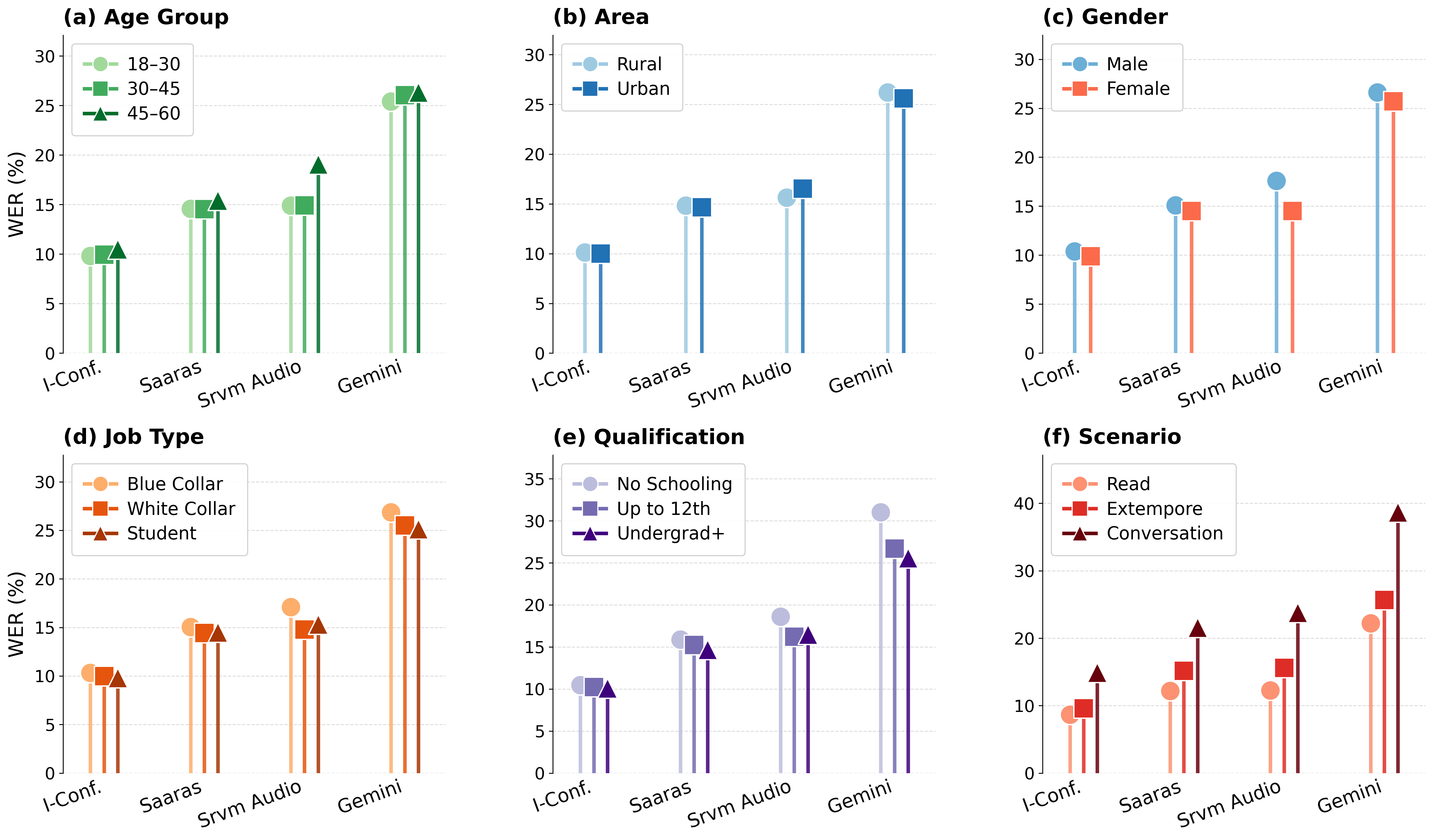}
    \caption{WER (\%) across speaker demographics and recording scenarios, averaged over 22 Indian languages.}
    \label{fig:demographics}
\end{figure}

\subsection{The State of ASR across regions of India}

In Figure \ref{fig:demographics}, district-level WER from I-Conf reveals substantial geographic variation. The lowest-WER districts are concentrated in West Bengal (Hooghly: 5.4, South 24 Parganas: 6.0), Gujarat (Bharuch: 4.6, Gandhinagar: 8.9), and Uttar Pradesh (Shahjahanpur: 7.7), while the highest cluster in northern Karnataka (Koppal: 54.9, Gadag: 52.2, Dharwad: 47.6) and Kerala (Kollam: 46.7, Kottayam: 45.3), broadly consistent with the greater phonological and morphological complexity of Dravidian languages. Bihar districts exhibit a wide internal spread (Gaya: 14.2 vs.\ Araria: 34.5, Begusarai: 35.0), suggesting dialectal fragmentation within Hindi-belt states independently contributes to recognition difficulty. Most strikingly, northern Karnataka deviates sharply from the south (Hassan: 18.2, Chikkaballapur: 23.3) by over 35 WER points, possibly reflecting Urdu and Marathi contact varieties on northern Kannada dialects, a challenge that aggregate language-level evaluations would obscure.


\subsection{Does duration and speaking rate affect performance?}

\begin{figure}[t]
    \centering
    \includegraphics[width=0.9\linewidth]{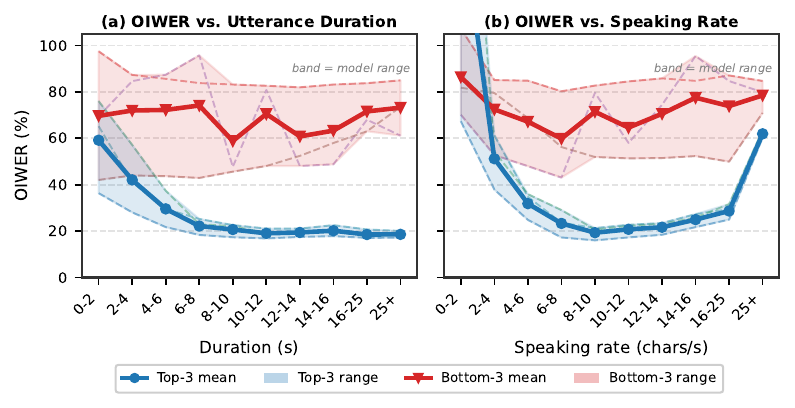}
    \caption{OIWER (\%) vs.\ utterance duration (a) and speaking rate (b) for top-3
and bottom-3 models.}
    \label{fig:duration_rate}
\end{figure}

As shown in Figure~\ref{fig:duration_rate}, top-performing models (I-Conf, Saaras, Sarvam Audio)
degrade sharply for short utterances ($<$4s) and recover to stable OIWER ($\sim$17--20\%) beyond
8s, while bottom-tier models remain uniformly poor across all conditions.
A symmetric U-shaped degradation appears for speaking rate, with optimal performance
at 8--12 chars/s and sharp increases at both extremes. Notably, Azure STT exhibits a monotonically \textit{increasing} OIWER with duration (42\%
at 0--2s vs.\ 73.5\% at 25+s), inverting the trend seen in all other systems.
At very slow speech rates (0--2 chars/s), even the best model (I-Conf) reaches
67.4\% OIWER, suggesting disfluent or heavily paused speech is a near-universal failure mode.

\subsection{Which acoustic events challenge ASR models most?}

To better understand model behavior across acoustic conditions, we group event labels into three categories based on average WER and inter-model range (max $-$ min WER across models).

\noindent\textbf{Universally Hard} audio events (avg.\ WER $> 38$, range $< 25$) are those where all models struggle consistently, suggesting the acoustic-linguistic mismatch is fundamental rather than model-specific. Events such as \textit{Choir}, \textit{Yodeling}, and \textit{Mantra} fall into this category, with \textit{Mantra} exhibiting the tightest inter-model spread (range $= 6.5$), indicating near-universal failure.

\noindent\textbf{Universally Easy} audio events (avg.\ WER $< 25$, range $< 15$) represent conditions where background acoustics pose little challenge to any model. Events like \textit{Writing}, \textit{Mechanical fan}, and \textit{Narration} achieve low WER across all systems, likely because they co-occur with minimally interfering speech.

\noindent\textbf{High Disagreement} audio events (range $> 50$) reveal stark model-specific vulnerabilities. \textit{Children shouting} (range $= 133.7$) sees Sarvam Audio produce a WER of 177 while others stay below 50, and \textit{Screaming} splits models into two clear camps: three models exceed 90 WER while the remaining three stay near 40. \textit{Telephone bell ringing} exposes the Saaras (WER $\approx 80\text{--}89$) against others at $\approx 22\text{--}28$, suggesting non-speech tonal sounds reveal training data gaps.



\section{Conclusion}

We introduced \textit{Vimarsha}, a benchmark for more faithful evaluation of ASR systems across the 22 scheduled Indian languages. It combines demographically diverse on-field recordings with challenging in-the-wild audio and introduces a lattice of variations that captures multiple valid transcriptions for each utterance. Evaluations of ten state-of-the-art ASR systems reveal substantial shifts in model rankings under realistic acoustic conditions, clear geographic and demographic performance disparities, and systematic failure modes across speaking rates and acoustic events.  We release the benchmark, evaluation protocol, and model results to support the development of ASR systems that better reflect real-world deployment conditions and the linguistic variability of Indian languages.

\section{Acknowledgements}

We thank Digital India Bhashini, MeitY, Government of India, for generously supporting this work. We are grateful to the EkStep Foundation and Nilekani Philanthropies for their generous support, which enabled the recruitment of human resources and provided access to the cloud infrastructure essential for carrying out this work. We thank the entire team of AI4Bharat, especially the language experts who contributed to benchmark construction and verification, thereby enabling the successful completion of this project. Finally, we acknowledge everyone who contributed to this initiative in any capacity and helped make this work meaningful and impactful.

\section{Generative AI Use Disclosure}
Generative AI tools were used solely for language polishing and editing during the preparation of this manuscript. These tools assisted with improving clarity, grammar, and conciseness of the writing. No generative AI system was used to generate experimental results, analyses, figures, or scientific conclusions. All technical content, experiments, and interpretations were developed and verified by the authors.






\bibliographystyle{IEEEtran}
\bibliography{references}

\end{document}